\documentclass[journal]{IEEEtran}
\usepackage{graphicx}
\usepackage{xcolor}
\usepackage{cite}
\begin{document}
%
% paper title
% Titles are generally capitalized except for words such as a, an, and, as,
% at, but, by, for, in, nor, of, on, or, the, to and up, which are usually
% not capitalized unless they are the first or last word of the title.
% Linebreaks \\ can be used within to get better formatting as desired.
% Do not put math or special symbols in the title.
\title{Dietary Assessment with Multimodal ChatGPT: A Systematic Analysis}

\author{Frank P.-W. Lo,
       Jianing~Qiu, Zeyu~Wang, Junhong~Chen, Bo~Xiao, Wu~Yuan,~\IEEEmembership{Senior Member,~IEEE}, Stamatia~Giannarou, Gary~Frost and~Benny~Lo,~\IEEEmembership{Senior Member,~IEEE}% <-this % stops a space
\thanks{This work was supported by the Innovative Passive Dietary Monitoring Project funded by the Bill \& Melinda Gates Foundation under Grant OPP1171395. F. P.-W. Lo, Z. Wang, J. Chen, B. Xiao, and S. Giannarou are with the Hamlyn Centre, Imperial College London, London SW7 2AZ, UK, {e-mail:\{po.lo15, zeyu.wang20, junhong.chen16, b.xiao, stamatia.giannarou\}@imperial.ac.uk}. J. Qiu and W. Yuan are with the Department of Biomedical Engineering, The Chinese University of Hong Kong, Hong Kong, {e-mail:\{jianingqiu, wyuan\}@cuhk.edu.hk}. G. Frost and B. Lo are with Department of Metabolism, Digestion and Reproduction, Faculty of Medicine, Imperial College London, London, SW7 2AZ, UK, {e-mail:\{g.frost, benny.lo\}@imperial.ac.uk}%
        }}

% The paper headers
\markboth{Journal of \LaTeX\ Class Files,~Vol.~14, No.~8, August~2015}%
{Shell \MakeLowercase{\textit{et al.}}: Bare Demo of IEEEtran.cls for IEEE Journals}
% The only time the second header will appear is for the odd numbered 

% make the title area
\maketitle

% As a general rule, do not put math, special symbols or citations
% in the abstract or keywords.
\begin{abstract}
Conventional approaches to dietary assessment are primarily grounded in self-reporting methods or structured interviews conducted under the supervision of dietitians. These methods, however, are often subjective, potentially inaccurate, and time-intensive. Although artificial intelligence (AI)-based solutions have been devised to automate the dietary assessment process, these prior AI methodologies encounter challenges in their ability to generalize across a diverse range of food types, dietary behaviors, and cultural contexts. This results in AI applications in the dietary field that possess a narrow specialization and limited accuracy. Recently, the emergence of multimodal foundation models such as GPT-4V powering the latest ChatGPT has exhibited transformative potential across a wide range of tasks (e.g., Scene understanding and image captioning) in numerous research domains. These models have demonstrated remarkable generalist intelligence and accuracy, capable of processing various data modalities. In this study, we explore the application of multimodal ChatGPT within the realm of dietary assessment. Our findings reveal that GPT-4V excels in food detection under challenging conditions with accuracy up to 87.5\% without any fine-tuning or adaptation using food-specific datasets. By guiding the model with specific language prompts (e.g., African cuisine), it shifts from recognizing common staples like rice and bread to accurately identifying regional dishes like banku and ugali. Another GPT-4V's standout feature is its contextual awareness. GPT-4V can leverage surrounding objects as scale references to deduce the portion sizes of food items, further enhancing its accuracy in translating food weight into nutritional content. This alignment with the USDA National Nutrient Database underscores GPT-4V's potential to advance nutritional science and dietary assessment techniques. 
% \textcolor{red}{you didn't mention prompts, without prompts, it is not multimodal!!!!! better mention results of utlising prompts to refine the xxx}
\end{abstract}

% Note that keywords are not normally used for peerreview papers.

\IEEEpeerreviewmaketitle

\section{Introduction}
\label{sect:intro}  % \label{} allows reference to this section

The traditional approach to dietary assessment typically relies on self-reports or interviews supervised by dietitians~\cite{shim2014dietary}, such as 24-Hour dietary recall (24HR), dietary record (DR), and food frequency questionnaires (FFQs). However, recent advancements in artificial intelligence (AI) have significantly reshaped the methodologies employed by dietitians for assessing dietary intake and influenced how the general public manages their dietary habits~\cite{lo2020image,konstantakopoulos2023review}. Over the past few years, there has been a marked increase in the development of AI algorithms and applications specifically designed for automatic food and ingredient recognition~\cite{bossard2014food,qiu2019mining,jiang2019multi,min2019ingredient,min2023large,chen2016deep,qiu2020counting,konstantakopoulos2023automated}, food segmentation~\cite{wu2021large}, food volume estimation~\cite{lo2018food,lo2019novel,lo2019point2volume,lo2022intelligent,lu2021partially,lu2020artificial}, and recipe retrieval~\cite{salvador2017learning,chen2017cross,chen2018deep,zhu2019r2gan,salvador2021revamping} or generation~\cite{salvador2019inverse,wang2020structure,wang2022learning}, as well as machine-learning-enabled smart sensors to detect eating events~\cite{dong2013detecting,shen2016assessing,saphala2022proximity} and behaviors~\cite{8606156,kyritsis2020data} in free-living settings. In addition to the algorithmic and hardware advancements, the swift progression of AI-based technological solutions for dietary assessment can be partly attributed to the curation of large-scale datasets by the community, for instance, Food2K~\cite{min2023large}, a dataset that encompasses over 1 million food images across 2,000 categories, and Recipe1M+~\cite{marin2021recipe1m+}, which comprises over 1 million recipes and 13 million food images. 

Despite these advancements, previous AI methodologies applied to dietary assessment often exhibit a limited scope, processing discrete tasks such as merely recognizing food items~\cite{bossard2014food,qiu2019mining,jiang2019multi,min2019ingredient,min2023large} or solely retrieving the recipe for a particular dish~\cite{salvador2017learning,chen2017cross,chen2018deep,zhu2019r2gan,salvador2021revamping}. These approaches lack a comprehensive understanding of dietary knowledge, common practices, assessment workflow, and the variations in eating behaviors and cultural contexts of individuals, hence falling short of achieving the overarching objective of accurately quantifying an individual's food and nutrient intake, followed by personalized advice on healthy diet and health management.

Recently, the emergence of AI foundation models~\cite{bommasani2021opportunities} has presented a new perspective. These models are characterized by their vast number of parameters, the extensive and diverse data utilized for training, and their remarkable versatility. They are capable of addressing a broad spectrum of tasks through post-training adaptation~\cite{bommasani2021opportunities}.  This ushers in novel paradigms for the design of methodologies across numerous research fields, and foundation models are in fact revolutionizing various sectors within healthcare~\cite{moor2023foundation,qiu2023large}, offering generalist intelligence and large-scale generalizability across multiple tasks such as the recent VisionFM~\cite{qiu2023visionfm} model that is able to process eight ophthalmic imaging modalities and solve a wide array of ophthalmic tasks. The advent of foundation models opens up new opportunities for enhancing AI-based dietary assessment. Their primary advantage lies in their ability to associate each individual case with a wealth of general and domain-specific knowledge they have acquired, thereby potentially providing a more precise dietary assessment, followed by personalized dietary and health advice.

\begin{figure*}[ht]
\centering
\includegraphics[width=\linewidth]{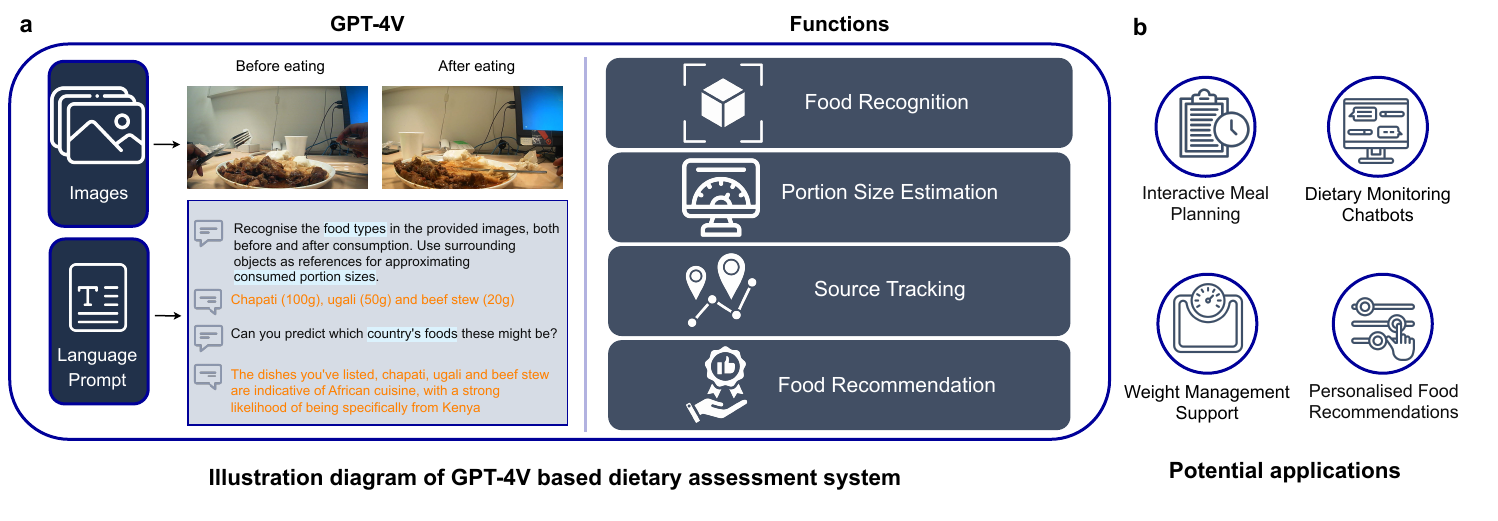}
\caption{(a) Schematic overview of the GPT-4V based dietary assessment system. The left section outlines the multimodal input methods, incorporating both visual (images before and after eating) and textual (language prompts) data. The right section details the primary functions of the system. (b) The potential applications of GPT-4V in the fields of nutrition and health.}
\label{fig:my_label} % Label for referencing the figure in text
\end{figure*}

\begin{figure*}[ht]
\centering
\includegraphics[width=0.95\linewidth]{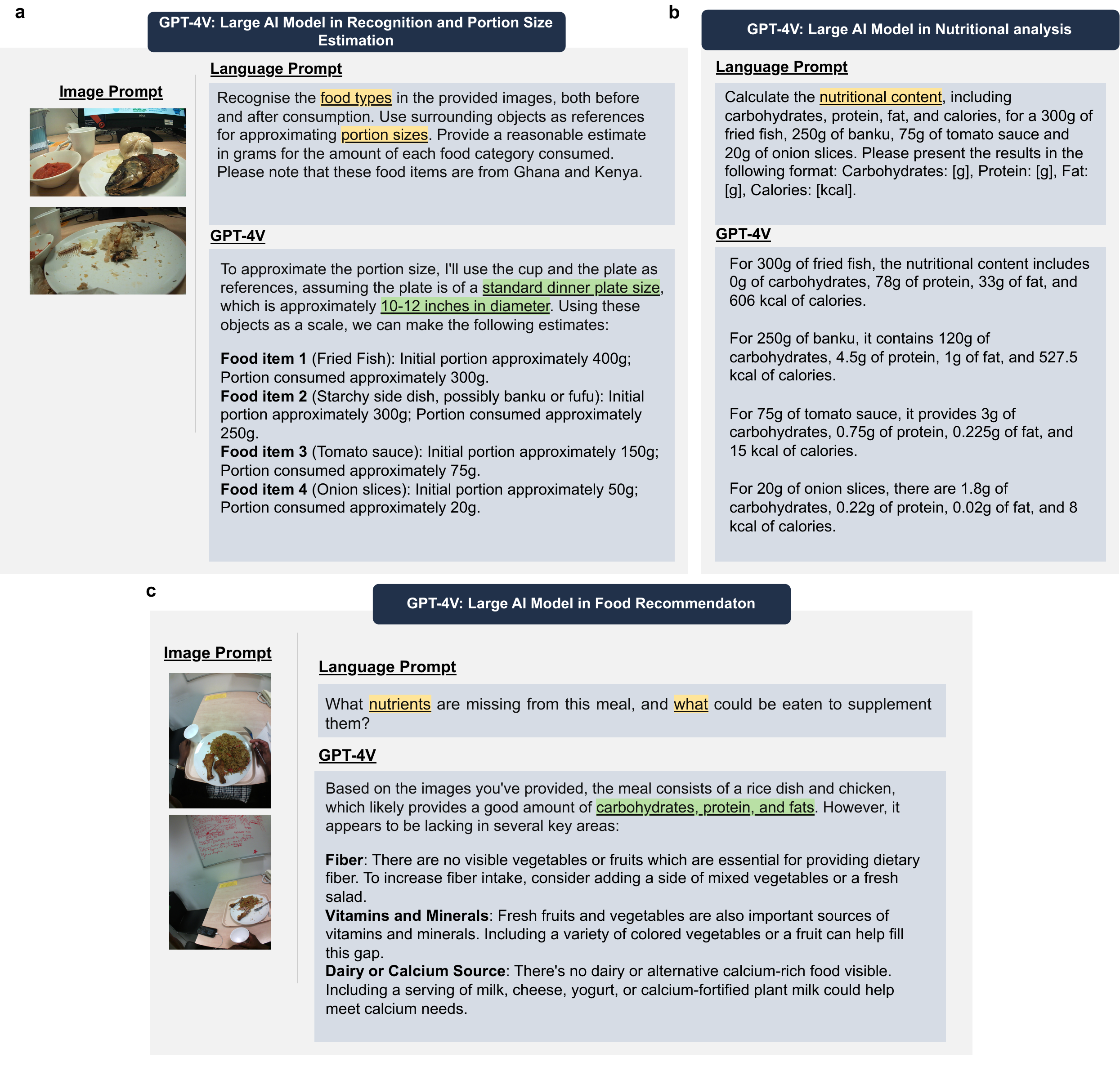}

\caption{\textbf{Integrative nutritional assessment by GPT-4V: displayed in this figure are the multifaceted food analysis capabilities of GPT-4V, focused on traditional dishes from Ghana and Kenya.} (a) GPT-4V first recognizes food items and estimates the consumed portion size using images taken before a meal and after consumption, utilizing common utensils as reference points for scale. (b) This figure highlights GPT-4V's proficiency in determining the precise nutrient content of the meal, providing a detailed breakdown of carbohydrates, proteins, fats, and calories. (c) The bottom section presents GPT-4V's nutritional critique and enhancement suggestions for the meal, addressing the nutrient adequacy and pinpointing deficiencies in dietary fiber, vitamins, minerals, and calcium, while recommending additional food items to achieve a nutritionally balanced diet. This illustration underscores GPT-4V's potential as a tool for dietary analysis and health promotion within diverse cultural contexts.}
\label{method_gpt} % Label for referencing the figure in text
\end{figure*}

In this study, we leverage one of seminal multimodal foundation models, i.e., GPT-4V~\cite{gpt4v_system_card}, which powers the latest generation of ChatGPT, for dietary assessment. Previously, GPT-4V has demonstrated impressive capabilities in understanding visual content, and responding to queries with a wide spectrum of knowledge and a high degree of natural language proficiency. Our investigation has yielded significant insights into the capabilities of GPT-4V in the realm of food detection and nutritional analysis.

Our key findings are as follows:
\begin{itemize}
\item GPT-4V demonstrates satisfactory performance in identifying food items, achieving an improved accuracy of 87.5\% when language prompts regarding the food’s origin are provided. We also discovered that GPT-4V possesses an impressive potential for assessing dietary intake from images captured in dark environments, a scenario where previous vision-based methods often find challenging, yet eating in low-light is common in some low- and middle-income countries (LMICs). This improved assessment accuracy can be attributed to the inherent reasoning abilities of GPT-4V as well as its well-established knowledge graph to complement visual processing, enabling it to recognize food items captured in low-light or in dark effectively with a few language prompts contextualizing the specific dietary and cultural background.

% \textcolor{red}{we should discuss this in the discussion that GPT-4V's ability to recognise food items in dark is hard to be grounded as to whether it really has improved visual processing skills, or it actually also cannot recognise through image, but can guess through language prompts and its learned wide knowledge about the culture, diet, etc}

\item GPT-4V's adeptness in estimating portion sizes is remarkable, with the ability to utilize the scale of nearby objects as a reference. This proficiency extends to logically deducing portion sizes at both pre- and post-consumption, where its estimations closely align with the Weighed Food Records (WFR), the standard ground truth in dietary assessment, and demonstrating comparable performance to human estimates, with GPT-4V's error at 54.6g and dietitians' at 43.6g benchmarked on a real-world dietary intake dataset~\cite{jobarteh2023evaluation}. This is particularly noteworthy given the complexity of the task, especially under conditions employing a monocular RGB camera without the existence of fiducial markers of a known size.

\item GPT-4V's ability to convert the weight of food items into their nutritional components shows a high level of accuracy, correlating closely with the nutritional data from the USDA National Nutrient Database\footnote{https://fdc.nal.usda.gov/index.html}. This highlights GPT-4V's potential for effective cross-referencing and accurate interpretation within the context of nutritional science.

% \textcolor{red}{need to present some highlight quantative results here}

% \textcolor{red}{we should summarize some quantitative results here. AND limitations. We need to stress that without prompts, GPT-4V will misrecognise banku as rice, indicating bias in its training data, this point is very important and I guarantee people will cite this paper just because of this point, i.e., bias in African data. Let's summarize our findings using bullet points here to make it more convenient for readers to grasp.}

% \item GPT-4V's recognition capabilities demonstrate certain constraints in the absence of explicit prompts. A notable instance of this is its misclassification of banku as rice, indicating an underlying bias within its training dataset. Nevertheless, when provided with specific language prompts that contextualise the food as belonging to African cuisines, the model corrects its identification accordingly.
\end{itemize}

\section{Related Work}
\label{sect:related_work}

Our study primarily investigates the vision-language feature of multimodal ChatGPT, i.e., the underlying GPT-4V model. Hence, this section mainly reviews preceding work in the field of dietary AI that employs both visual and textual dietary data. 

Previously, the use of deep learning-based image captioning, a multimodal technique, has been proposed in the context of dietary intake monitoring and assessment~\cite{qiu2023egocentric}. This approach allows for the simultaneous identification of food types, estimation of portion sizes for food items, and discernment of the eating scenario - such as communal eating or solitary dining. Additionally, it incorporates the estimation of food container volume to accurately quantify a subject's dietary intake in uncontrolled, real-world scenarios. Liu et al.~\cite{liu2018multi} devised a system which can jointly analyze a dish photo, the description, and user information to rate the healthiness of the dish, and subsequently provide suggestions on improving the diet of the user. Multimodal AI techniques have also been used to develop food recipe retrieval systems~\cite{salvador2017learning,chen2017cross,chen2018deep,zhu2019r2gan,salvador2021revamping}, which are primarily based on learning a common embedding space for food image-recipe pairs, aligning textual data of recipes with visual content of food images. Furthermore, AI researchers have developed systems that can generate a recipe for a dish based on its captured image~\cite{salvador2019inverse,wang2020structure,wang2022learning}. Such systems commonly leverage a hierarchical or structured way of generating recipes for food images, for example, by first predicting the ingredients and then inferring the cooking instructions~\cite{salvador2019inverse}.

AI foundation models like GPT-4V use self-supervised learning to learn task-agnostic representations from large-scale data points. Peng et al.~\cite{peng2022clustering} previously investigated the use of self-supervised learning to mine discriminative representations from large quantity of egocentric dietary intake images, and demonstrated that the learned image representations can facilitate eating-related event identification.

Since the introduction of GPT-4V, there has been some work studying its effectiveness in biomedicine and healthcare areas, for example, in medical image analysis~\cite{yang2023dawn,wu2023can}. Early findings suggest that while it is a great leap towards more intelligent multimodal foundation models that can approach a wider range of tasks, GPT-4V still face challenges such as hallucination~\cite{cui2023holistic}, and inconsistent performance during multi-round dialogue especially requiring domain knowledge such as medical diagnosis~\cite{wu2023can}. 

% Benchmark latest GPT-4V \textcolor{blue}{we better insert them in proper context rather than explicitly mentioning them, eg previous work on GPT-4V xxx xxx xxx}: 
% \begin{enumerate}
%     \item Holistic Analysis of Hallucination in GPT-4V(ision): Bias and Interference Challenges~\cite{cui2023holistic}
%     \item Medical imaging~\cite{wu2023can}
%     \item Microsoft~\cite{yang2023dawn}
% \end{enumerate}

\section{Detailed Information and Methods}

\subsection{Data Collection}
This study aims to analyze the effectiveness of GPT-4V in conducting nutritional analysis from images taken by an egocentric camera during passive dietary intake monitoring. We used dietary intake data of adults of Kenyan and Ghanaian origin living in London, collected following our study protocol~\cite{jobarteh2020development}. Statistical data analysis and the demographics of subjects can be found in~\cite{jobarteh2023evaluation}. Participants were assigned either an eButton~\cite{sun2015exploratory} or an AIM~\cite{doulah2020automatic} camera to passively record their dietary intake. A standardized Salter Brecknell scale was used to initially weigh the food items, which were then served to participants. The study deviated from traditional methods by not requiring participants to consume all the food; instead, they were instructed to eat until they were full. The remaining food was then weighed again, and the difference between the initial and final weights was recorded as the actual consumed food weight, also known as Weighed Food Records (WFR). Experienced dietitians also participated to estimate the weight of the food visually as human accuracy in estimating food weights. Their assessments were compared with the results from GTP-4V to evaluate its performance. To ensure fairness and accuracy, the assessors received training in recognizing Ghanaian and Kenyan foods and in the technique of visual food weight estimation. 

\subsection{Multimodal Foundation Model - GPT-4V }

GPT-4V, a cutting-edge multimodal language model, demonstrates its proficiency in processing both image and text inputs to generate highly skilled outputs. Built upon a transformer-based neural network architecture, GPT-4V has showcased human-level performance across various professional and academic benchmarks. While specific technical details of GPT-4V are not yet disclosed by OpenAI, its status as a state-of-the-art Large Multimodal Model (LMM) is pivotal. Its application in dietary assessment raises intriguing questions about the potential for LMMs to achieve human-level performance in this domain, promising significant advancements in the field.

\subsubsection{\textbf{Food Detection and Recognition}}
The challenges inherent in existing food recognition systems primarily revolve around the diversity of food types. Supervised learning approaches struggle to cover the vast majority of possible foods, as manual labeling and training data for each food item can be impractical. However, this limitation is where self-supervised learning emerges as a pivotal solution for the future of AI-based dietary assessment. Self-supervised learning models, like GPT-4V, excel in handling diverse and unstructured data. They are not reliant on explicit labeling but instead learn from extensive datasets, including a wide range of content. GPT-4V's proficiency is underpinned by its extensive training on vast amounts of data from diverse sources, enabling it to acquire a deep and broad understanding of various topics, including dietary information. In the initial phase of the methodology, GPT-4V was tasked with food detection and recognition within the provided images. This involved analyzing two separate images: one depicting the meal before consumption and the other after consumption. 

\subsubsection{\textbf{Portion Size Estimation}}

 To enhance the accuracy of dietary assessment, we instructed GPT-4V to effectively utilize contextual information from the images, including the relative sizes and positions of food items. GPT-4V's contextual understanding enabled the model to estimate the portion sizes of each food item more accurately. Additionally, we employed various language prompts to test the outcomes of the portion size estimation. The responses generated by GPT-4V were structured according to the format detailed in the language prompt. This structured data format was crucial for subsequent nutritional analysis. To ensure the robustness and reliability of the estimated portion sizes, the methodology incorporated a confidence interval determination step. This step involved repeating the portion size estimation process 4 times that captured the variations of the model's estimations. The language prompt used for food recognition and portion size estimation is shown in Figure \ref{method_gpt}(a).

\subsubsection{\textbf{Nutritional Analysis}}

The next phase of the methodology focused on calculating the nutritional content of specific food items. Using a distinct language prompt, shown in Figure \ref{method_gpt}(b), GPT-4V was instructed to compute key nutritional parameters, including carbohydrates, protein, fat, and calories, for the identified food items. The outputs were compared with the ground truth (GT), derived from WFR and the nutritional data from the USDA National Nutrient Database.

\subsubsection{\textbf{Nutrient Gap Identification and Food Recommendations}}

Following the retrieval of nutritional data, the methodology extended its inquiry to identify potential nutrient gaps within the analyzed meal. GPT-4V was prompted to identify which essential nutrients might be lacking in the meal composition. Moreover, the model was asked to provide recommendations for supplementary foods that could be consumed to address these identified nutrient deficiencies. This aspect of the methodology aimed to offer practical dietary guidance by suggesting appropriate foods to complement the analyzed meal and enhance its overall nutritional completeness.

\begin{figure}[tb]
\centering
\includegraphics[width=0.9\linewidth]{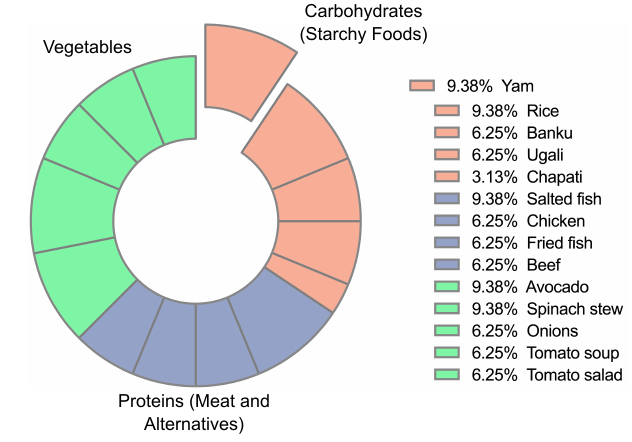}
\caption{\textbf{Food distribution in the study.} This figure illustrates the breakdown of food items utilized in the study, categorized into vegetables, carbohydrates (starchy foods), and proteins (meat and alternatives).}
\label{food_distribution} % Label for referencing the figure in text
\end{figure}

\section{Experimental Results}

\begin{table}[tb]
\centering
\caption{Quantitative Results of Food Detection using GPT-4V}
\vspace{5pt}

\label{food_detection_metric}
\resizebox{\columnwidth}{!}{%
\begin{tabular}{c|cccc}
\hline
 & Accuracy & Precision & Recall & F1 score \\ \hline
\begin{tabular}[c]{@{}c@{}}Language\\ Prompt\end{tabular} & 0.719 & 0.793 & 0.885 & 0.836 \\ \hline
\begin{tabular}[c]{@{}c@{}}Language Prompt \\ with Contextual Information\end{tabular} & \textbf{0.875} & \textbf{0.903} & \textbf{0.966} & \textbf{0.933} \\ \hline
\end{tabular}%
}
\end{table}

\subsection{Food Detection using GPT-4V}

In this study, we first present a quantitative analysis of food detection capabilities using GPT-4V, with a focus on understanding the influence of language prompts on the food identification accuracy. Figure \ref{food_distribution} illustrates the food distribution used in the study. There are 14 African food items, which can be categorized into three groups including vegetables, carbohydrates (starchy foods), and proteins (meat and alternatives). The quantitative results, as shown in Table \ref{food_detection_metric}, outline the accuracy, precision, recall, and F1 score of the food detection by GPT-4V. In scenarios with language prompts that did not include information about the origin of food, the model's accuracy was 71.9\%, with a precision of 79.3\%, a recall of 88.5\%, and an F1 score of 83.6\%. In contrast, when language prompts were enriched with contextual information about the food's origin, there was a notable increase in the results of food detection. The model achieved an accuracy of up to  87.5\%, precision of 90.3\%, recall of 96.6\%, and the F1 score of 93.3\%. This comparison reveals a pattern: when the origin of the food is not specified, the GPT-4V model frequently misclassified white starchy foods as rice. However, when prompts specified the food's Kenyan and Ghanian origin, the model accurately identified the food items as banku or ugali, which are traditional African foods. These findings indicate that language prompts containing contextual cues about food origin can significantly enhance the LMMs' ability to accurately identify culturally specific foods.

\subsection{Portion Size Estimation using GPT-4V}

\begin{table*}[htb]
\centering
\caption{Absolute Error in Nutrient Estimation Across Varied Eating Episodes}
\label{episode}
\resizebox{\linewidth}{!}{%
\begin{tabular}{cccccccc}
\begin{tabular}[c]{@{}c@{}}Episode \\ \#\end{tabular}  & \begin{tabular}[c]{@{}c@{}}No. of \\ Food Items\end{tabular}  & Food Lists & \begin{tabular}[c]{@{}c@{}}Absolute Error of \\ Consumed Portion Size (g)\end{tabular} & \begin{tabular}[c]{@{}c@{}}Absolute Error of \\ Carbohydrate (g)\end{tabular} & \begin{tabular}[c]{@{}c@{}}Absolute Error of\\ Protein (g)\end{tabular} & \begin{tabular}[c]{@{}c@{}}Absolute Error of\\ Fat (g)\end{tabular} & \begin{tabular}[c]{@{}c@{}}Absolute Error of \\ Calorie (kcal)\end{tabular} \\ \hline
1 & 4 & \begin{tabular}[c]{@{}c@{}}Yam, avocado, \\ spinach stew, salted fish\end{tabular} & 136 & 12.4 & 2.3 & 7.9 & 94.4 \\ \hline
2 & 2 & Fried rice, chicken drumstick & 25 & 12.6 & 1.4 & 5.6 & 51.0 \\ \hline
3 & 2 & Fried rice, chicken drumstick & 102 & 46.2 & 8.3 & 6.4 & 6.0 \\ \hline
4 & 4 & \begin{tabular}[c]{@{}c@{}}Onions, tilapia fish, \\ tomato soup, banku\end{tabular} & 18 & 36.5 & 2.0 & 16.5 & 110.7 \\ \hline
5 & 4 & \begin{tabular}[c]{@{}c@{}}Avocado, salted fish, \\ yam, spinach stew\end{tabular} & 70 & 5.5 & 2.8 & 10.6 & 99.7 \\ \hline
6 & 3 & \begin{tabular}[c]{@{}c@{}}Tomato salad, roasted beef, \\ ugali\end{tabular} & 204 & 5.2 & 39.2 & 29.6 & 390.8 \\ \hline
7 & 2 & Tomato salad, fried rice & 44 & 36.8 & 5.3 & 8.8 & 124.2 \\ \hline
8 & 3 & Chapati, ugali, beef stew & 382 & 99.3 & 40.1 & 12.4 & 628.0 \\ \hline
9 & 4 & \begin{tabular}[c]{@{}c@{}}Avocado, yam, \\ spinach stew, salted fish\end{tabular} & 71 & 8.0 & 11.3 & 5.9 & 5.7 \\ \hline
10 & 4 & \begin{tabular}[c]{@{}c@{}}Tilapia fish, onions, \\ tomato soup, banku\end{tabular} & 31 & 68.7 & 27.1 & 12.3 & 1.2 \\ \hline
\end{tabular}%
}
\end{table*}

\begin{figure}[tb]
\centering
\includegraphics[width=\linewidth]{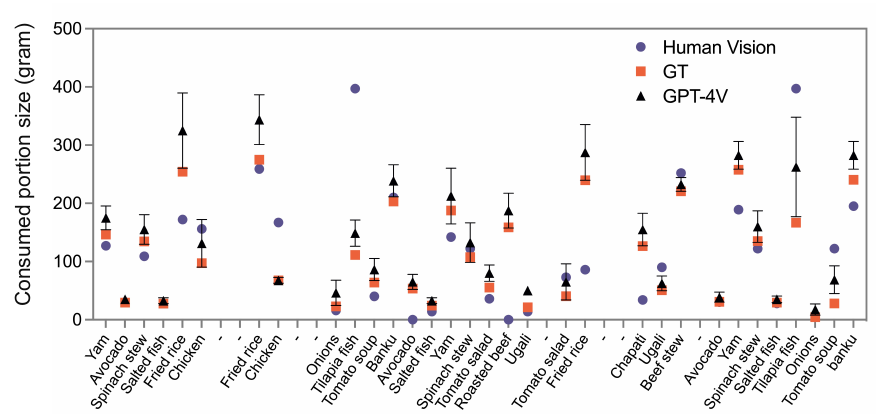}
\caption{\textbf{Comparison of consumed portion size estimations by different methods.} This figure presents a side-by-side comparison of portion size estimations for various food items. Each food item, represented in 10 eating episodes within this study, is plotted to illustrate the variance and accuracy of each method's estimate.}
% \textcolor{red}{for the legend, it is better to arrange in the order of GPT-4V, Human, GT}
\label{consume_portionsize} % Label for referencing the figure in text
\end{figure}

\begin{figure}[htb]
\centering
\includegraphics[width=0.9\columnwidth]{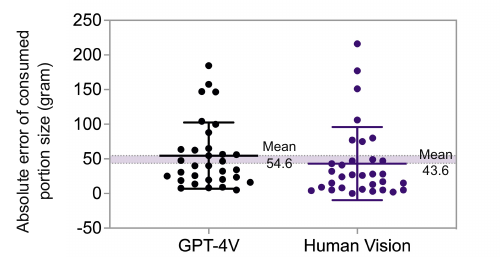}
\caption{\textbf{Comparative analysis of GPT-4V and human estimates in portion size estimation.} It demonstrates the absolute error in consumed portion size estimation between GPT-4V and human vision. The mean absolute error for GPT-4V is 54.6g compared to 43.6g from human estimates, showcasing GPT-4V's comparable performance to human-level accuracy in the task of portion size estimation}
\label{AE_portionsize} % Label for referencing the figure in text
\end{figure}

In Figure \ref{consume_portionsize}, we present a comparative analysis of consumed portion size estimations derived from three methods including the GPT-4V, human vision, and the ground truth. The data shows a variety of food items, each subjected to 10 unique eating episodes within the study (See Table \ref{episode} for the details of each episode). Besides, the error bars, plotted for the GPT-4V estimations, are based on four separate trials, providing insights into the model's reliability. Remarkably, these trials display minimal variation, indicating the estimations are consistently close to one another. The trend observed in the plot reveals that the GPT-4V's predictions generally align closely with the ground truth, showing the model's potential for accurate portion size estimation. In Figure \ref{AE_portionsize}, we further conduct a comparison between GPT-4V and human visual assessments, utilizing the metric of Absolute Error (AE). The plot distinctly visualizes individual data points representing the absolute error for each estimation compared to the ground truth. The GPT-4V's estimations have a mean absolute error of 54.6g. In comparison, human estimates reflect a slightly superior accuracy, with a lower mean absolute error of 43.6g. This comparative analysis reveals that the performance of GPT-4V aligns closely with that of human vision, illustrating the GPT-4V's capability in approximating human-like estimation. 

\begin{figure*}[htb]
\centering
\includegraphics[width=\textwidth]{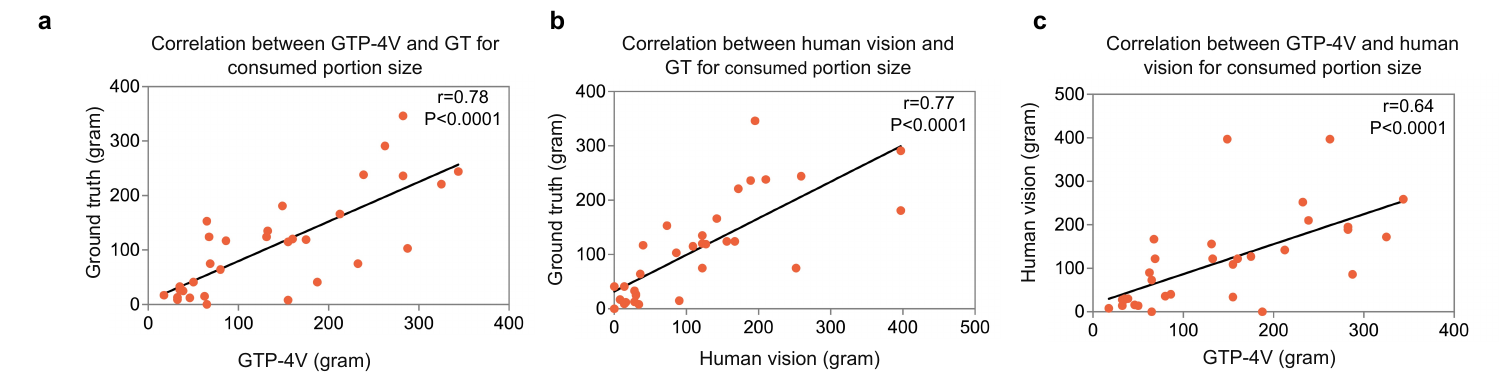}
\caption{\textbf{Comparative analysis of GTP-4V's predictions and human estimates against ground truth for consumed portion size estimation.} (a) Correlation plot for GTP-4V's predictions vs. ground truth in portion size, with a high correlation coefficient (r = 0.78, P \textless 0.0001), suggesting the model's effectiveness in this specific task of portion estimation (b) Correlation plot for human estimates vs. ground truth in portion size, showing strong agreement (r = 0.77, P \textless 0.0001). (c) Correlation plot comparing GTP-4V's predictions with human estimates (r = 0.64, P \textless 0.0001).}
\label{correlation} % Label for referencing the figure in text
\end{figure*}

\begin{figure}[htb]
\centering
\includegraphics[width=0.9\columnwidth]{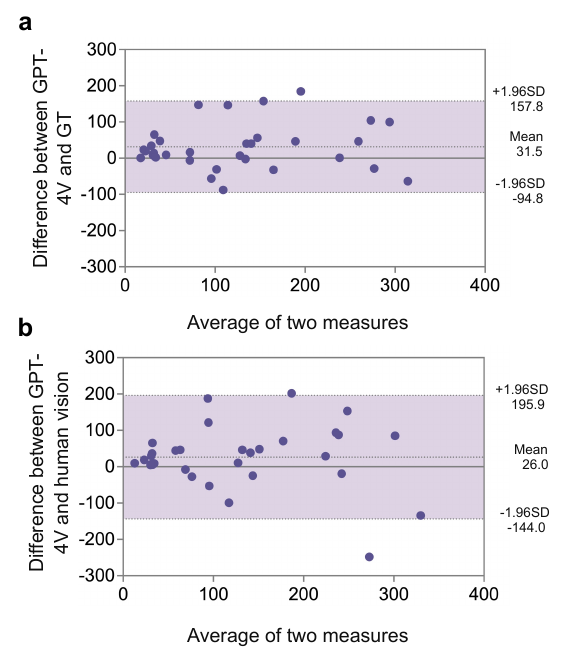}
\caption{\textbf{Bland-Altman analyses.} (a) This Bland-Altman plot compares the GPT-4V model to GT (b) This Bland-Altman plot compares GPT-4V to human vision. The vast majority of data points fall within the limits of agreement in these two plots, showcasing that there is a good agreement between the methods.}
\label{bland} % Label for referencing the figure in text
\end{figure}

\begin{figure*}[ht]
\centering
\includegraphics[width=\textwidth]{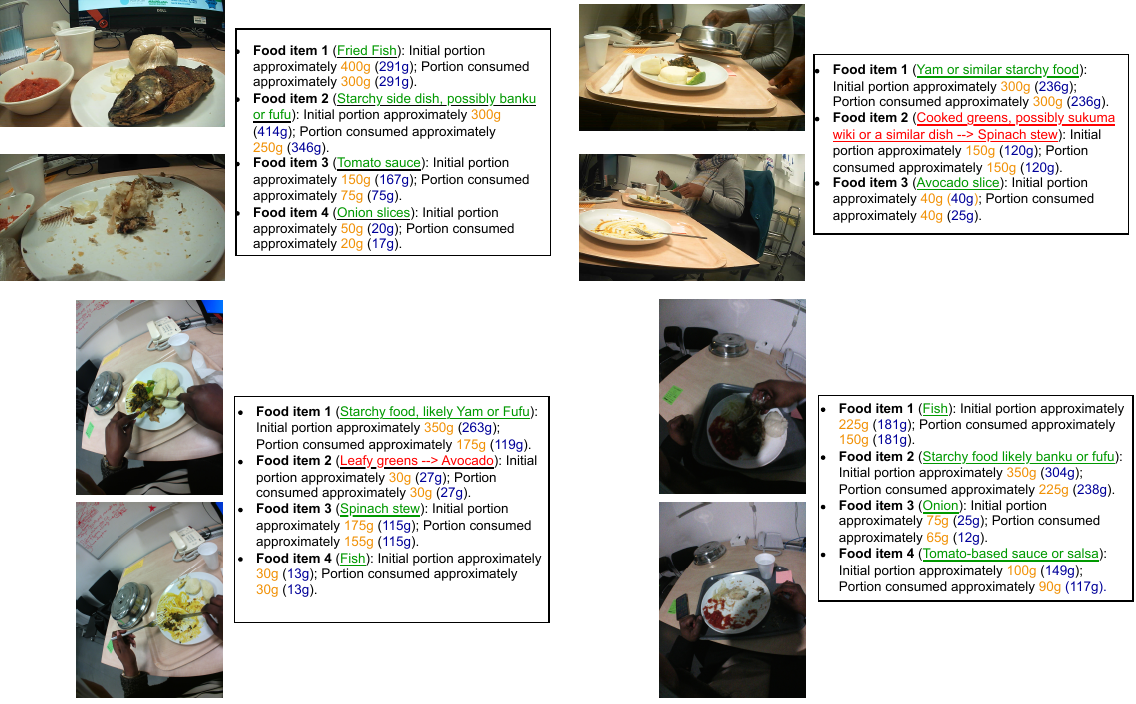}
\caption{\textbf{Qualitative results of sample dietary intake.} This figure presents a selection of meals with itemized descriptions, highlighting the initial portion sizes and the amounts consumed by study participants. Green labels indicate correctly recognized food items, while red labels point to errors. The orange label shows the estimated value. The blue label at the end of each description specifies the actual portion size, serving as the ground truth.}
\label{quality} % Label for referencing the figure in text
\end{figure*}

In Figure \ref{correlation}(a), in the analysis to determine the correlation between the GPT-4V and ground truth for estimating consumed portion sizes, we employed Pearson's correlation coefficient as a measure of the strength and direction of the association between the two variables. The analysis yielded a Pearson r value of 0.78, indicating a strong positive correlation between the estimates provided by GPT-4V and those obtained through GT. This result is statistically significant with a p-value of less than 0.0001. Further statistical analysis also showed that the 95\% confidence interval for the Pearson r extends from 0.5845 to 0.8847, reinforcing the robustness of the correlation between the GPT-4V method and GT. For Figure \ref{correlation}(b), a Pearson's correlation analysis was also conducted to assess the relationship between human vision estimates and the GT for consumed portion sizes. The resulting Pearson r value of 0.77 signifies a strong positive correlation, almost paralleling the correlation observed with the GPT-4V. In Figure \ref{correlation}(c), the correlation between GPT-4V and human vision estimates yields an r value of 0.64. This correlation analysis demonstrates the potential of the GPT-4V, showing again its ability to understand and process portion size estimation in a manner similar to humans. Although a slightly higher correlation does not necessarily mean that the GPT-4V is superior, it does indicate that the GPT-4V is a promising tool for this specific task, especially in scenarios that require rapid and large-scale dietary assessment. Last, we conducted the Bland-Altman analysis to evaluate the GPT-4V model. In Figure \ref{bland}(a), the Bland-Altman plot for the GPT-4V method vs. GT confirms the model's effectiveness, demonstrating strong agreement in predictions of consumed portion sizes in grams. Another Bland-Altman plot, as shown in Figure \ref{bland}(b), comparing GPT-4V to human vision also indicates that GPT-4V is consistent with human assessments, showcasing GPT-4V's comparable performance to human-level accuracy. Qualitative results of sample dietary intake can also be found in Figure \ref{quality}.

\subsection{The Impact of Varied Language Prompts on Portion Size Estimation}

We further provided a comparative analysis of the Mean Absolute Error (MAE) in portion size estimations when employing the GPT-4V with two distinct language prompts (See Table \ref{reference_standard}). We first instructed the model to use surrounding objects as scale references, and then we asked the model to rely more on prior knowledge, e.g., relying on the standard serving size of each food item. The first method, based on reference objects, yielded an MAE of 52.9g with a standard deviation (SD) of 51.3g. The second approach, utilizing standard serving sizes, resulted in a slightly higher MAE of 57.2g with an SD of 49.4g. This finding is pivotal, indicating that guiding the GPT-4V to utilize scale references can facilitate portion size estimation. 

\begin{table}[tb]
\centering
\caption{Comparative Analysis of Portion Size Estimation Using Reference Objects and Standard Serving Sizes}
\label{reference_standard}
\resizebox{\columnwidth}{!}{%
\begin{tabular}{c|ccc}
\hline
 & MAE & SD & Confidence Interval (95\%) \\ \hline
Reference Objects-based & \textbf{52.9} & 51.3 & 52.9$\pm$17.8 \\
Standard Serving Sizes-based & 57.2 & 49.4 & 57.2$\pm$17.1 \\ \hline
\end{tabular}%
}
\end{table}

\subsection{Nutritional Analysis using GPT-4V}

\begin{figure}[t]
\centering
\includegraphics[width=\columnwidth]{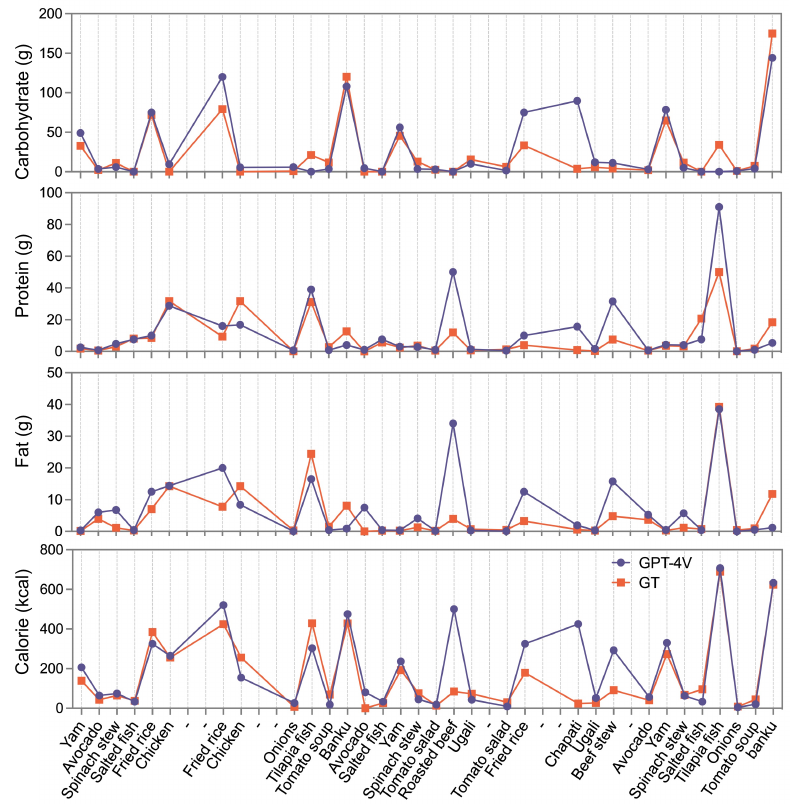}
\caption{\textbf{Nutrient estimation for various food items.} This figure shows the GPT-4V model's predictions for carbohydrate, protein, fat, and calorie content of various food items compared with GT.}
\label{nutrient_variousitems} % Label for referencing the figure in text
\end{figure}

After portion size estimation, GPT-4V model was utilized to convert food weights into specific nutrient data, focusing on key dietary components: carbohydrates, protein, fat, and calories. While GPT-4V is capable of a broader range of nutritional analysis, these nutrients were selected for their relevance to the objectives of this study. The results obtained were compared with the ground truth, which was derived from WFR in conjunction with nutritional data from the USDA National Nutrient Database. This comparison is crucial for evaluating the accuracy of the GPT-4V model's nutrient estimation capabilities against a well-established nutritional standard. Table \ref{episode} shows the absolute error in estimated nutrient content across 10 different eating episodes. Each episode consists of a variety of food items, and the nutritional content (carbohydrates, protein, fat, and calories) estimated by the GPT-4V model has been aggregated for each episode. The finding suggests that the accuracy of nutrient estimation is highly dependent on the accuracy of the portion size estimation. For instance, episodes 6 and 8 show some deviation in portion size estimation from the GT, which in turn significantly impacted the accuracy of the nutritional estimates. In other eating episodes where the portion size estimation was close to the actual value, the nutrients estimation tends to align close to their corresponding nutritional value (i.e., calorie estimation error tends to be less than 100kcal for each episode when portion size is predicted correctly). In Table \ref{aggregate_table}, we further present a comparative analysis of nutritional estimation errors from two different perspectives: episode-based and food item-based. The episode-based analysis considers the aggregate errors from the 10 episodes mentioned earlier, while the food item-based analysis evaluates data from 32 individual food items featured across those episodes. For each perspective, the table lists the Mean Absolute Error (MAE) with a 95\% Confidence Interval (CI) and Standard Deviation (SD) for four nutritional components: carbohydrate, protein, fat, and calorie content. Nutrient estimation for each individual food item can be found in Figure \ref{nutrient_variousitems}.

\begin{table*}[tb]
\centering
\caption{Comparative Analysis of Nutritional Estimation Errors: Episode-Based vs. Food Item-Based Approaches}
\label{aggregate_table}
\resizebox{0.9\textwidth}{!}{%
\begin{tabular}{cccccccccc}
\hline
 & \begin{tabular}[c]{@{}c@{}}Sample \\ size\end{tabular} & \begin{tabular}[c]{@{}c@{}}MAE of carbohydrate,\\  CI (95\%) (g)\end{tabular} & SD  & \begin{tabular}[c]{@{}c@{}}MAE of protein,\\  CI (95\%) (g)\end{tabular} & SD & \begin{tabular}[c]{@{}c@{}}MAE of fat,\\ CI (95\%) (g)\end{tabular} & SD & \begin{tabular}[c]{@{}c@{}}MAE of calorie,\\ CI (95\%) (kcal)\end{tabular} & SD \\ \hline
\begin{tabular}[c]{@{}c@{}}Food item-based\end{tabular} & 32 & 12.3$\pm$6.1 & 17.7 & 6.5$\pm$3.6 & 10.4 & 4.1$\pm$2.1 & 6.1 & 69.2$\pm$34.7 & 100.2 \\
\begin{tabular}[c]{@{}c@{}}Episode-based\end{tabular} & 10 & 33.1$\pm$19.4 & 31.4 & 14.0$\pm$9.6 & 15.5 & 11.6$\pm$4.5 & 7.2 & 151.2$\pm$125.3 & 202.1 \\ \hline
\end{tabular}%
}
\end{table*}

% \begin{figure}[ht]
% \centering
% \includegraphics[width=\linewidth]{images/chatgpt_food_try3.pdf}

% \caption{This figure illustrates the process where GPT-4V utilizes a pair of images, one taken before and one after eating, to identify food items. Subsequently, it applies prior knowledge, likely concerning food density, along with environmental context, such as scale references, to accurately estimate the quantity of food consumed. \textcolor{red}{I think you can merge this figure with figure 1. You can put this figure under GPT-4V section of figure 1 to illustrate how GPT-4V works}}
% \label{fig:my_label4} % Label for referencing the figure in text
% \end{figure}

\section{Discussion}

\begin{figure}[tb]
\centering
\includegraphics[width=0.9\linewidth]{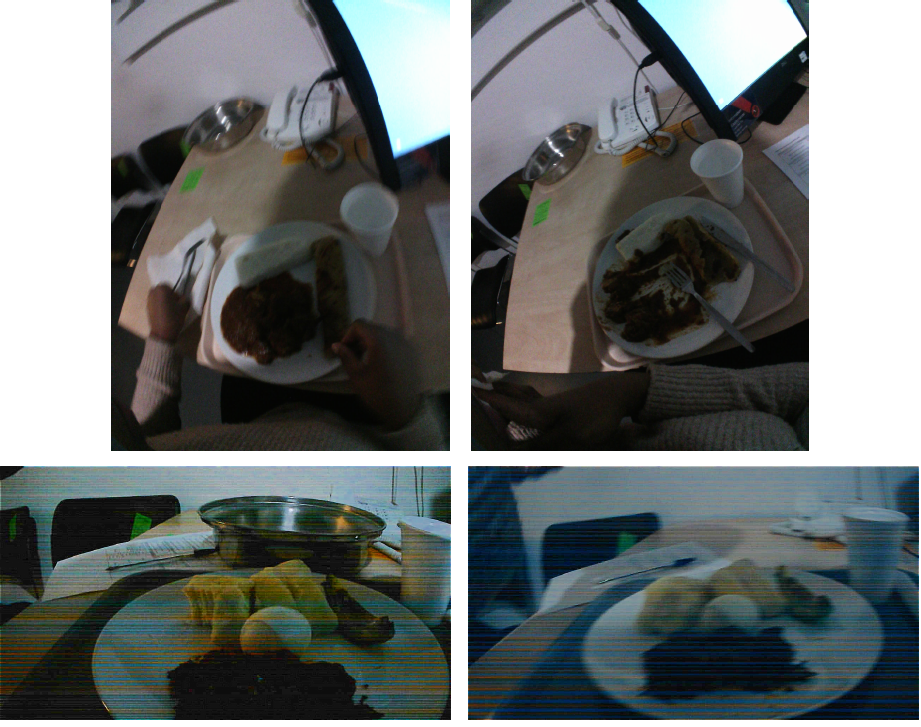}
\caption{\textbf{Images of food samples captured in low-light conditions.} In a dimly lit setting, the absence of contextual language prompts may lead GPT-4V to mis-recognize foods, often defaulting to common items such as rice and bread. However, when provided with language prompts that offer foods' origin, GPT-4V's performance in identifying the correct food items is improved, correctly recognizing items like banku and ugali.}
\label{dark} % Label for referencing the figure in text
\end{figure}

In this research, we analyzed the potential of multimodal foundation models in dietary assessment, with a particular focus on GPT-4V. We initially discovered that GPT-4V demonstrates exceptional capabilities in food detection, even without fine-tuning or adaptation using a food-specific dataset. In addition, the GPT-4V model's ability to handle ambiguous food items is particularly noteworthy. When provided with contextual cues in the language prompts, such as the geographical origin of a food item, the model can utilize its extensive knowledge base to refine its recognition. This shows a sophisticated level of contextual understanding embedded in the model, allowing it to utilize external information effectively to enhance its recognition accuracy. Moreover, our research highlighted the model's capability to handle images with low visibility, such as those captured in dark environment, as shown in Figure \ref{dark}. In scenarios lacking of guidance, the model may default to identifying more commonly known food items, such as rice and bread. However, upon receiving specific language prompts (e.g., African cuisine), the GPT-4V model demonstrates a remarkable ability to adjust its deduction and accurately identify regional food items, like banku.

Although GPT-4V has certain capabilities, we also noticed that it is not specifically designed to understand complex spatial relationships or perform precise spatial analysis tasks. For instance, when estimating the volume or portion size of food items, precise spatial perception is required to identify and calculate the 3D attributes of objects, which typically exceeds the capabilities of GPT-4V, and it may not be mature enough to accurately estimate the portion size. In our experiments, we observed that consumed food items weighing less than 30g could not be easily estimated by the model. Nevertheless, it is still capable of achieving a close human-level accuracy in the task of portion size estimation.

While this study is mainly focused on solitary eating, eating is sometimes a social activity, and in many cultures and countries such as African and Asian countries, communal eating and food sharing are very common. Food sharing scenarios require analyzing the entire eating episode to quantify each individual dietary intake~\cite{qiu2019assessing,lei2021assessing}. While GPT-4V can understand short video clips, its video analysis ability needs to be improved so that long dietary intake episodes can be analyzed to estimate food and nutrient intake in food sharing or communal eating scenarios. Furthermore, the Be My Eye application~\cite{bemyeyes} which is powered by GPT-4V to assist visually impaired or blind people as proxy eyes may inspire the integration or connection of foundation models like GPT-4V with dietary intake monitoring and assessment systems~\cite{sun2015exploratory,doulah2020automatic}, providing a new level of on-node intelligence and facilitating pervasive dietary intake monitoring and assessment.

Nevertheless, the use of AI foundation models in dietary assessment faces other obstacles. Inadequate or biased data can result in inaccurate or biased results. For instance, without providing contextual cues, GPT-4V tends to recognize typical African food like banku as rice or bread. In addition, the complexity of multimodal foundation models can make them difficult to understand and interpret, which may lead to mistrust, misuse, and resistance among users. There are also potential risks and ethical considerations associated with using AI foundation models in dietary assessment. One major risk is the potential misuse of dietary data and adversarially prompting foundation models, which may result in revealing sensitive information about an individual's health status or sensitive personal information~\cite{qiu2023egocentric}, especially those inside the training dataset. There is also the risk of inaccurate dietary recommendations due to errors or biases in the AI algorithms as well as in the data used to develop them, which could have negative health implications. Furthermore, there is a need for transparency in how the underlying mechanism of AI foundation models work and make decisions about dietary assessment, as well as safeguards to ensure the privacy, accuracy, and security of dietary data.

% \textcolor{red}{The limitation of this work?}

% \textcolor{red}{We need to share some of our insights into how future dietary assessment might be changed by generalist AI foundation models like ChatGPT} \textcolor{red}{e.g., the latest multimodal chatgpt can also read PDF files. We should say a bit about this functionality in discussion}

\section{Future Directions}

To achieve better performance in dietary assessment, accuracy in portion size estimation is crucial. To mitigate portion size estimation errors by GPT-4V, another way worth exploring in future work and may produce more robust dietary intake assessment is to estimate the bite size and count the number of bites~\cite{qiu2020counting} as previous study has revealed the correlation between bite counts and dietary intake~\cite{dong2012new}. Additionally, there are several potential approaches to enhancing the spatial perception abilities of large multimodal models (LMMs) so as to improve the accuracy in estimating portion sizes. One such method involves the use of depth image. Data from depth cameras, which offer detailed information about the distances and dimensions of object items, can be used to train LMMs. This integration of depth information would provide the model with a richer understanding of an object's position and size within a given space, significantly improving its ability to estimate portion sizes accurately. Another approach is the utilization of 3D simulation data/physics engines. Training LMMs with data derived from these environments could lead to a substantial improvement in their comprehension of 3D spaces. Such environments offer a more realistic representation of objects and their spatial relationships, which could be invaluable for the model's learning process and subsequent application in real-world scenarios. Considering the current capabilities of GPT models to interpret PDF content, a future enhancement in the field of dietary assessment could see dietitians incorporating health standards or Standard Operating Procedures (SOPs) into these documents. GPT models could leverage this information to conduct dietary assessments that are both standardized according to professional guidelines and customized to individual requirements. This would take AI-based dietary assessment to a new level of precision and personalization, closely aligning it with the nuanced demands of nutritional counseling and health management.

\section{Conclusion}

In conclusion, this study underscores the potential of GPT-4V, a multimodal foundation model powering the latest ChatGPT, in transforming the field of dietary assessment. It demonstrates a high accuracy rate of up to 87.5\% in food detection under various conditions without specific fine-tuning. Its ability to accurately identify a wide range of foods, from common staples to regional dishes like banku and ugali, by utilizing specific language prompts, is particularly noteworthy. Additionally, GPT-4V's contextual awareness allows it to use surrounding objects as a scale reference, thereby enhancing its precision in estimating portion sizes and converting these into nutritional content. This capability highlights GPT-4V's immense potential to revolutionize nutritional science and dietary assessment methodologies.

\bibliographystyle{unsrt}
\bibliography{reference}

\begin{thebibliography}{10}

\bibitem{shim2014dietary}
Jee-Seon Shim, Kyungwon Oh, and Hyeon~Chang Kim.
\newblock Dietary assessment methods in epidemiologic studies.
\newblock {\em Epidemiology and health}, 36, 2014.

\bibitem{lo2020image}
Frank P-W Lo, Yingnan Sun, Jianing Qiu, and Benny Lo.
\newblock Image-based food classification and volume estimation for dietary assessment: A review.
\newblock {\em IEEE journal of biomedical and health informatics}, 24(7):1926--1939, 2020.

\bibitem{konstantakopoulos2023review}
Fotios~S Konstantakopoulos, Eleni~I Georga, and Dimitrios~I Fotiadis.
\newblock A review of image-based food recognition and volume estimation artificial intelligence systems.
\newblock {\em IEEE Reviews in Biomedical Engineering}, 2023.

\bibitem{bossard2014food}
Lukas Bossard, Matthieu Guillaumin, and Luc Van~Gool.
\newblock Food-101--mining discriminative components with random forests.
\newblock In {\em Computer Vision--ECCV 2014: 13th European Conference, Zurich, Switzerland, September 6-12, 2014, Proceedings, Part VI 13}, pages 446--461. Springer, 2014.

\bibitem{qiu2019mining}
Jianing Qiu, Frank P-W Lo, Yingnan Sun, Siyao Wang, and Benny Lo.
\newblock Mining discriminative food regions for accurate food recognition.
\newblock In {\em British Machine Vision Conference (BMVC)}, 2019.

\bibitem{jiang2019multi}
Shuqiang Jiang, Weiqing Min, Linhu Liu, and Zhengdong Luo.
\newblock Multi-scale multi-view deep feature aggregation for food recognition.
\newblock {\em IEEE Transactions on Image Processing}, 29:265--276, 2019.

\bibitem{min2019ingredient}
Weiqing Min, Linhu Liu, Zhengdong Luo, and Shuqiang Jiang.
\newblock Ingredient-guided cascaded multi-attention network for food recognition.
\newblock In {\em Proceedings of the 27th ACM International Conference on Multimedia}, pages 1331--1339, 2019.

\bibitem{min2023large}
Weiqing Min, Zhiling Wang, Yuxin Liu, Mengjiang Luo, Liping Kang, Xiaoming Wei, Xiaolin Wei, and Shuqiang Jiang.
\newblock Large scale visual food recognition.
\newblock {\em IEEE Transactions on Pattern Analysis and Machine Intelligence}, 2023.

\bibitem{chen2016deep}
Jingjing Chen and Chong-Wah Ngo.
\newblock Deep-based ingredient recognition for cooking recipe retrieval.
\newblock In {\em Proceedings of the 24th ACM international conference on Multimedia}, pages 32--41, 2016.

\bibitem{qiu2020counting}
Jianing Qiu, Frank P-W Lo, Shuo Jiang, Ya-Yen Tsai, Yingnan Sun, and Benny Lo.
\newblock Counting bites and recognizing consumed food from videos for passive dietary monitoring.
\newblock {\em IEEE Journal of Biomedical and Health Informatics}, 25(5):1471--1482, 2020.

\bibitem{konstantakopoulos2023automated}
Fotios~S Konstantakopoulos, Eleni~I Georga, and Dimitrios~I Fotiadis.
\newblock An automated image-based dietary assessment system for mediterranean foods.
\newblock {\em IEEE Open Journal of Engineering in Medicine and Biology}, 2023.

\bibitem{wu2021large}
Xiongwei Wu, Xin Fu, Ying Liu, Ee-Peng Lim, Steven~CH Hoi, and Qianru Sun.
\newblock A large-scale benchmark for food image segmentation.
\newblock In {\em Proceedings of the 29th ACM International Conference on Multimedia}, pages 506--515, 2021.

\bibitem{lo2018food}
Frank P-W Lo, Yingnan Sun, Jianing Qiu, and Benny Lo.
\newblock Food volume estimation based on deep learning view synthesis from a single depth map.
\newblock {\em Nutrients}, 10(12):2005, 2018.

\bibitem{lo2019novel}
Frank P-W Lo, Yingnan Sun, Jianing Qiu, and Benny Lo.
\newblock A novel vision-based approach for dietary assessment using deep learning view synthesis.
\newblock In {\em 2019 IEEE 16th International Conference on Wearable and Implantable Body Sensor Networks (BSN)}, pages 1--4. IEEE, 2019.

\bibitem{lo2019point2volume}
Frank P-W Lo, Yingnan Sun, Jianing Qiu, and Benny~PL Lo.
\newblock Point2volume: A vision-based dietary assessment approach using view synthesis.
\newblock {\em IEEE Transactions on Industrial Informatics}, 16(1):577--586, 2019.

\bibitem{lo2022intelligent}
Frank P-W Lo, Yao Guo, Yingnan Sun, Jianing Qiu, and Benny Lo.
\newblock An intelligent vision-based nutritional assessment method for handheld food items.
\newblock {\em IEEE Transactions on Multimedia}, 2022.

\bibitem{lu2021partially}
Ya~Lu, Thomai Stathopoulou, and Stavroula Mougiakakou.
\newblock Partially supervised multi-task network for single-view dietary assessment.
\newblock In {\em 2020 25th International Conference on Pattern Recognition (ICPR)}, pages 8156--8163. IEEE, 2021.

\bibitem{lu2020artificial}
Ya~Lu, Thomai Stathopoulou, Maria~F Vasiloglou, Stergios Christodoulidis, Zeno Stanga, and Stavroula Mougiakakou.
\newblock An artificial intelligence-based system to assess nutrient intake for hospitalised patients.
\newblock {\em IEEE transactions on multimedia}, 23:1136--1147, 2020.

\bibitem{salvador2017learning}
Amaia Salvador, Nicholas Hynes, Yusuf Aytar, Javier Marin, Ferda Ofli, Ingmar Weber, and Antonio Torralba.
\newblock Learning cross-modal embeddings for cooking recipes and food images.
\newblock In {\em Proceedings of the IEEE conference on computer vision and pattern recognition}, pages 3020--3028, 2017.

\bibitem{chen2017cross}
Jing-jing Chen, Chong-Wah Ngo, and Tat-Seng Chua.
\newblock Cross-modal recipe retrieval with rich food attributes.
\newblock In {\em Proceedings of the 25th ACM international conference on Multimedia}, pages 1771--1779, 2017.

\bibitem{chen2018deep}
Jing-Jing Chen, Chong-Wah Ngo, Fu-Li Feng, and Tat-Seng Chua.
\newblock Deep understanding of cooking procedure for cross-modal recipe retrieval.
\newblock In {\em Proceedings of the 26th ACM international conference on Multimedia}, pages 1020--1028, 2018.

\bibitem{zhu2019r2gan}
Bin Zhu, Chong-Wah Ngo, Jingjing Chen, and Yanbin Hao.
\newblock R2gan: Cross-modal recipe retrieval with generative adversarial network.
\newblock In {\em Proceedings of the IEEE/CVF Conference on Computer Vision and Pattern Recognition}, pages 11477--11486, 2019.

\bibitem{salvador2021revamping}
Amaia Salvador, Erhan Gundogdu, Loris Bazzani, and Michael Donoser.
\newblock Revamping cross-modal recipe retrieval with hierarchical transformers and self-supervised learning.
\newblock In {\em Proceedings of the IEEE/CVF Conference on Computer Vision and Pattern Recognition}, pages 15475--15484, 2021.

\bibitem{salvador2019inverse}
Amaia Salvador, Michal Drozdzal, Xavier Gir{\'o}-i Nieto, and Adriana Romero.
\newblock Inverse cooking: Recipe generation from food images.
\newblock In {\em Proceedings of the IEEE/CVF Conference on Computer Vision and Pattern Recognition}, pages 10453--10462, 2019.

\bibitem{wang2020structure}
Hao Wang, Guosheng Lin, Steven~CH Hoi, and Chunyan Miao.
\newblock Structure-aware generation network for recipe generation from images.
\newblock In {\em Computer Vision--ECCV 2020: 16th European Conference, Glasgow, UK, August 23--28, 2020, Proceedings, Part XXVII 16}, pages 359--374. Springer, 2020.

\bibitem{wang2022learning}
Hao Wang, Guosheng Lin, Steven~CH Hoi, and Chunyan Miao.
\newblock Learning structural representations for recipe generation and food retrieval.
\newblock {\em IEEE Transactions on Pattern Analysis and Machine Intelligence}, 45(3):3363--3377, 2022.

\bibitem{dong2013detecting}
Yujie Dong, Jenna Scisco, Mike Wilson, Eric Muth, and Adam Hoover.
\newblock Detecting periods of eating during free-living by tracking wrist motion.
\newblock {\em IEEE journal of biomedical and health informatics}, 18(4):1253--1260, 2013.

\bibitem{shen2016assessing}
Yiru Shen, James Salley, Eric Muth, and Adam Hoover.
\newblock Assessing the accuracy of a wrist motion tracking method for counting bites across demographic and food variables.
\newblock {\em IEEE journal of biomedical and health informatics}, 21(3):599--606, 2016.

\bibitem{saphala2022proximity}
Addythia Saphala, Rui Zhang, and Oliver Amft.
\newblock Proximity-based eating event detection in smart eyeglasses with expert and data models.
\newblock In {\em Proceedings of the 2022 ACM International Symposium on Wearable Computers}, pages 59--63, 2022.

\bibitem{8606156}
Konstantinos Kyritsis, Christos Diou, and Anastasios Delopoulos.
\newblock Modeling wrist micromovements to measure in-meal eating behavior from inertial sensor data.
\newblock {\em IEEE Journal of Biomedical and Health Informatics}, 23(6):2325--2334, 2019.

\bibitem{kyritsis2020data}
Konstantinos Kyritsis, Christos Diou, and Anastasios Delopoulos.
\newblock A data driven end-to-end approach for in-the-wild monitoring of eating behavior using smartwatches.
\newblock {\em IEEE Journal of Biomedical and Health Informatics}, 25(1):22--34, 2020.

\bibitem{marin2021recipe1m+}
Javier Mar{\i}n, Aritro Biswas, Ferda Ofli, Nicholas Hynes, Amaia Salvador, Yusuf Aytar, Ingmar Weber, and Antonio Torralba.
\newblock Recipe1m+: A dataset for learning cross-modal embeddings for cooking recipes and food images.
\newblock {\em IEEE Transactions on Pattern Analysis and Machine Intelligence}, 43(1):187--203, 2021.

\bibitem{bommasani2021opportunities}
Rishi Bommasani, Drew~A Hudson, Ehsan Adeli, Russ Altman, Simran Arora, Sydney von Arx, Michael~S Bernstein, Jeannette Bohg, Antoine Bosselut, Emma Brunskill, et~al.
\newblock On the opportunities and risks of foundation models.
\newblock {\em arXiv preprint arXiv:2108.07258}, 2021.

\bibitem{moor2023foundation}
Michael Moor, Oishi Banerjee, Zahra Shakeri~Hossein Abad, Harlan~M Krumholz, Jure Leskovec, Eric~J Topol, and Pranav Rajpurkar.
\newblock Foundation models for generalist medical artificial intelligence.
\newblock {\em Nature}, 616(7956):259--265, 2023.

\bibitem{qiu2023large}
Jianing Qiu, Lin Li, Jiankai Sun, Jiachuan Peng, Peilun Shi, Ruiyang Zhang, Yinzhao Dong, Kyle Lam, Frank P-W Lo, Bo~Xiao, et~al.
\newblock Large ai models in health informatics: Applications, challenges, and the future.
\newblock {\em IEEE Journal of Biomedical and Health Informatics}, 2023.

\bibitem{qiu2023visionfm}
Jianing Qiu, Jian Wu, Hao Wei, Peilun Shi, Minqing Zhang, Yunyun Sun, Lin Li, Hanruo Liu, Hongyi Liu, Simeng Hou, et~al.
\newblock Visionfm: a multi-modal multi-task vision foundation model for generalist ophthalmic artificial intelligence.
\newblock {\em arXiv preprint arXiv:2310.04992}, 2023.

\bibitem{gpt4v_system_card}
OpenAI.
\newblock Gpt-4v(ision) system card.
\newblock 2023.

\bibitem{jobarteh2023evaluation}
Modou~L Jobarteh, Megan~A McCrory, Benny Lo, Konstantinos~K Triantafyllidis, Jianing Qiu, Jennifer~P Griffin, Edward Sazonov, Mingui Sun, Wenyan Jia, Tom Baranowski, et~al.
\newblock Evaluation of acceptability, functionality, and validity of a passive image-based dietary intake assessment method in adults and children of ghanaian and kenyan origin living in london, uk.
\newblock {\em Nutrients}, 15(18):4075, 2023.

\bibitem{qiu2023egocentric}
Jianing Qiu, Frank P-W Lo, Xiao Gu, Modou~L Jobarteh, Wenyan Jia, Tom Baranowski, Matilda Steiner-Asiedu, Alex~K Anderson, Megan~A McCrory, Edward Sazonov, et~al.
\newblock Egocentric image captioning for privacy-preserved passive dietary intake monitoring.
\newblock {\em IEEE Transactions on Cybernetics}, 2023.

\bibitem{liu2018multi}
Qi~Liu, Yue Zhang, Zhenguang Liu, Ye~Yuan, Li~Cheng, and Roger Zimmermann.
\newblock Multi-modal multi-task learning for automatic dietary assessment.
\newblock In {\em Proceedings of the AAAI Conference on Artificial Intelligence}, volume~32, 2018.

\bibitem{peng2022clustering}
Jiachuan Peng, Peilun Shi, Jianing Qiu, Xinwei Ju, Frank P-W Lo, Xiao Gu, Wenyan Jia, Tom Baranowski, Matilda Steiner-Asiedu, Alex~K Anderson, et~al.
\newblock Clustering egocentric images in passive dietary monitoring with self-supervised learning.
\newblock In {\em 2022 IEEE-EMBS International Conference on Biomedical and Health Informatics (BHI)}, pages 01--04. IEEE, 2022.

\bibitem{yang2023dawn}
Zhengyuan Yang, Linjie Li, Kevin Lin, Jianfeng Wang, Chung-Ching Lin, Zicheng Liu, and Lijuan Wang.
\newblock The dawn of lmms: Preliminary explorations with gpt-4v (ision).
\newblock {\em arXiv preprint arXiv:2309.17421}, 9, 2023.

\bibitem{wu2023can}
Chaoyi Wu, Jiayu Lei, Qiaoyu Zheng, Weike Zhao, Weixiong Lin, Xiaoman Zhang, Xiao Zhou, Ziheng Zhao, Ya~Zhang, Yanfeng Wang, et~al.
\newblock Can gpt-4v (ision) serve medical applications? case studies on gpt-4v for multimodal medical diagnosis.
\newblock {\em arXiv preprint arXiv:2310.09909}, 2023.

\bibitem{cui2023holistic}
Chenhang Cui, Yiyang Zhou, Xinyu Yang, Shirley Wu, Linjun Zhang, James Zou, and Huaxiu Yao.
\newblock Holistic analysis of hallucination in gpt-4v (ision): Bias and interference challenges.
\newblock {\em arXiv preprint arXiv:2311.03287}, 2023.

\bibitem{jobarteh2020development}
Modou~L Jobarteh, Megan~A McCrory, Benny Lo, Mingui Sun, Edward Sazonov, Alex~K Anderson, Wenyan Jia, Kathryn Maitland, Jianing Qiu, Matilda Steiner-Asiedu, et~al.
\newblock Development and validation of an objective, passive dietary assessment method for estimating food and nutrient intake in households in low-and middle-income countries: A study protocol.
\newblock {\em Current developments in nutrition}, 4(2):nzaa020, 2020.

\bibitem{sun2015exploratory}
Mingui Sun, Lora~E Burke, Thomas Baranowski, John~D Fernstrom, Hong Zhang, Hsin-Chen Chen, Yicheng Bai, Yuecheng Li, Chengliu Li, Yaofeng Yue, et~al.
\newblock An exploratory study on a chest-worn computer for evaluation of diet, physical activity and lifestyle.
\newblock {\em Journal of healthcare engineering}, 6:1--22, 2015.

\bibitem{doulah2020automatic}
Abul Doulah, Tonmoy Ghosh, Delwar Hossain, Masudul~H Imtiaz, and Edward Sazonov.
\newblock “automatic ingestion monitor version 2”--a novel wearable device for automatic food intake detection and passive capture of food images.
\newblock {\em IEEE journal of biomedical and health informatics}, 25(2):568--576, 2020.

\bibitem{qiu2019assessing}
Jianing Qiu, Frank P-W Lo, and Benny Lo.
\newblock Assessing individual dietary intake in food sharing scenarios with a 360 camera and deep learning.
\newblock In {\em 2019 IEEE 16th International Conference on Wearable and Implantable Body Sensor Networks (BSN)}, pages 1--4. IEEE, 2019.

\bibitem{lei2021assessing}
Jiabao Lei, Jianing Qiu, Frank P-W Lo, and Benny Lo.
\newblock Assessing individual dietary intake in food sharing scenarios with food and human pose detection.
\newblock In {\em Pattern Recognition. ICPR International Workshops and Challenges: Virtual Event, January 10--15, 2021, Proceedings, Part V}, pages 549--557. Springer, 2021.

\bibitem{bemyeyes}
OpenAI.
\newblock Be my eyes.
\newblock 2023.

\bibitem{dong2012new}
Yujie Dong, Adam Hoover, Jenna Scisco, and Eric Muth.
\newblock A new method for measuring meal intake in humans via automated wrist motion tracking.
\newblock {\em Applied psychophysiology and biofeedback}, 37:205--215, 2012.

\end{thebibliography}

\end{document}